\documentclass{article}
\usepackage{spconf,amsmath,graphicx}
\usepackage{booktabs}
\usepackage{pbox}
\usepackage{enumitem}
\usepackage{adjustbox}
\setlist{nosep, leftmargin=14pt}
\usepackage{amssymb}  
\usepackage{graphicx} 
\usepackage{pifont}   
\usepackage{mwe} 
\usepackage{lipsum}
\usepackage{array} 
\usepackage{graphicx} 
\newcommand{\xmark}{\ding{55}} 
\usepackage{multirow}
\usepackage{multicol}
\usepackage{subcaption}

\title{CLISC: Bridging CLIP and SAM by Enhanced CAM for Unsupervised Brain Tumor Segmentation}
\name{Xiaochuan Ma$^{1}$, Jia Fu$^{2}$, Wenjun Liao$^{3}$, Shichuan Zhang$^{3}$, Guotai Wang$^{2,4}$\thanks{Corresponding author: Guotai Wang (guotai.wang@uestc.edu.cn).}}
\address{$^{1}$ School of Computer Science and Engineering, 
$^{2}$ School of Mechanical and Electrical Engineering, \\ University of Electronic Science and Technology of China, Chengdu, China \\
$^{3}$  Department of Radiation Oncology, Sichuan Cancer Hospital \& Institute, Sichuan Cancer Center,\\School of Medicine, University of Electronic Science and Technology of China, Chengdu, China\\
$^{4}$ Shanghai Artificial Intelligence Laboratory, Shanghai, China \\ }
\begin{document}
%
\maketitle
\begin{abstract}
Brain tumor segmentation is important for diagnosis of the tumor, and current deep-learning methods rely on a large set of annotated images for training, with high annotation costs. Unsupervised segmentation is promising to avoid human annotations while the performance is often limited. In this study, we present a novel unsupervised segmentation approach that leverages the capabilities of foundation models, and it consists of three main steps: \textbf{\textit{(1)}} A vision-language model (i.e., CLIP) is employed to obtain image-level pseudo-labels for training a classification network. Class Activation Mapping (CAM) is then employed to extract Regions of Interest (ROIs), where an adaptive masking-based data augmentation is used to enhance ROI identification. \textbf{\textit{(2)}} The ROIs are used to generate bounding box and point prompts for the Segment Anything Model (SAM) to obtain segmentation pseudo-labels. \textit{\textbf{(3)}} A 3D segmentation network is trained with the SAM-derived pseudo-labels, where low-quality pseudo-labels are filtered out in a self-learning process based on the similarity between the SAM's output and the network's prediction. Evaluation on the BraTS2020 dataset demonstrates that our approach obtained an average Dice Similarity Score (DSC) of 85.60\%, outperforming five state-of-the-art unsupervised segmentation methods by more than 10 percentage points. Besides, our approach outperforms directly using SAM for zero-shot inference, and its performance is close to fully supervised learning.

\end{abstract}

\section{Introduction}
\label{sec:intro}
 Magnetic Resonance Imaging (MRI) plays an important role in brain tumor diagnosis and treatment planning, due to its good soft-tissue contrast for visualizing tumor morphology and structure. Deep learning has achieved remarkable performance on brain tumor segmentation when trained with a large set of annotated images~\cite{wang2019automatic, jyothi2023deep, liu2022self}. However, the process of pixel-level annotation is labor-intensive and time-consuming. To deal with this problem, unsupervised segmentation is appealing for this task as it avoids manual annotation, leading to zero annotation cost.

Traditionally, researchers have used classic unsupervised learning methods such as clustering~\cite{abdel2015brain} and level-set~\cite{thapaliya2013level} for brain tumor segmentation. Their performance is often limited due to the use of low-level features and complex hyperparameter tuning. In recent years, unsupervised deep learning models have been introduced for this task with better performance due to their automatic feature learning ability from a large set of images. The main strategies include anomaly detection~\cite{baur2020bayesian, nguyen2021unsupervised, silva2022constrained} and tumor simulation~\cite{zhang2023self}. Anomaly detection typically relies on identifying regions that differ from normal samples, which can hardly obtain high performance in images with low contrasts and ambiguous boundaries. Tumor simulation methods often leverage prior knowledge of the shape and textures of tumors to generate synthetic tumor images for supervised learning, where the distribution shift from synthetic to real tumor images will limit their performance.

Recently, with the development of foundation models, the Segment Anything Model (SAM)~\cite{kirillov2023segment} has achieved state-of-the-art performance on various downstream segmentation tasks without requiring annotated images for a specific dataset for training. It employs prompts like bounding boxes and points for generalizable zero-shot inference, therefore has the potential for the segmentation of brain tumors. However, the quality of prompts highly affects the performance of SAM~\cite{kirillov2023segment}, and the prompts are often given by human, which still needs some manual efforts. In addition, the everything mode of SAM~\cite{kirillov2023segment} does not require human efforts, but the output is class agnostic, which cannot directly obtain brain tumor regions. Though some works~\cite{aleem2024test} have proposed to use vision-language models such as CLIP~\cite{radford2021learning} to get the label of regions generated by SAM~\cite{kirillov2023segment} for unsupervised segmentation, the zero-shot inference performance is limited due to the gap between natural images on which the CLIP/SAM was trained and medical images.

In this paper, we propose a novel unsupervised brain tumor segmentation method by adapting foundation models (CLIP and SAM) rather than directly using them for inference. The contribution is three-fold. First, we propose a framework named CLISC that bridges CLIP and SAM by enhanced Class Activation Maps (CAM) for unsupervised brain tumor segmentation, where image-level labels obtained by CLIP is used to train a classification model that obtains CAM, and the CAM  is used to generate prompts for SAM to obtain pseudo segmentation labels.  Second, to obtain high-quality prompts, we propose an Adaptive Masking-based Data Augmentation (AMDA) strategy for improved CAM quality. Thirdly, to reject low-quality segmentation pseudo-labels, we propose a SAM-Seg Similarity-based Filtering (S3F) strategy in a self-learning method for training a segmentation model. Evaluation on the BraTS2020 dataset shows that our method outperforms five state-of-the-art unsupervised segmentation methods by more than 10 percentage points with an average DSC of 85.60\%. Besides, our approach outperforms zero-shot SAM inference, achieving performance on par with fully supervised learning.

\begin{figure}[t]
  \centering
  \centerline{\includegraphics[width=8.5cm]{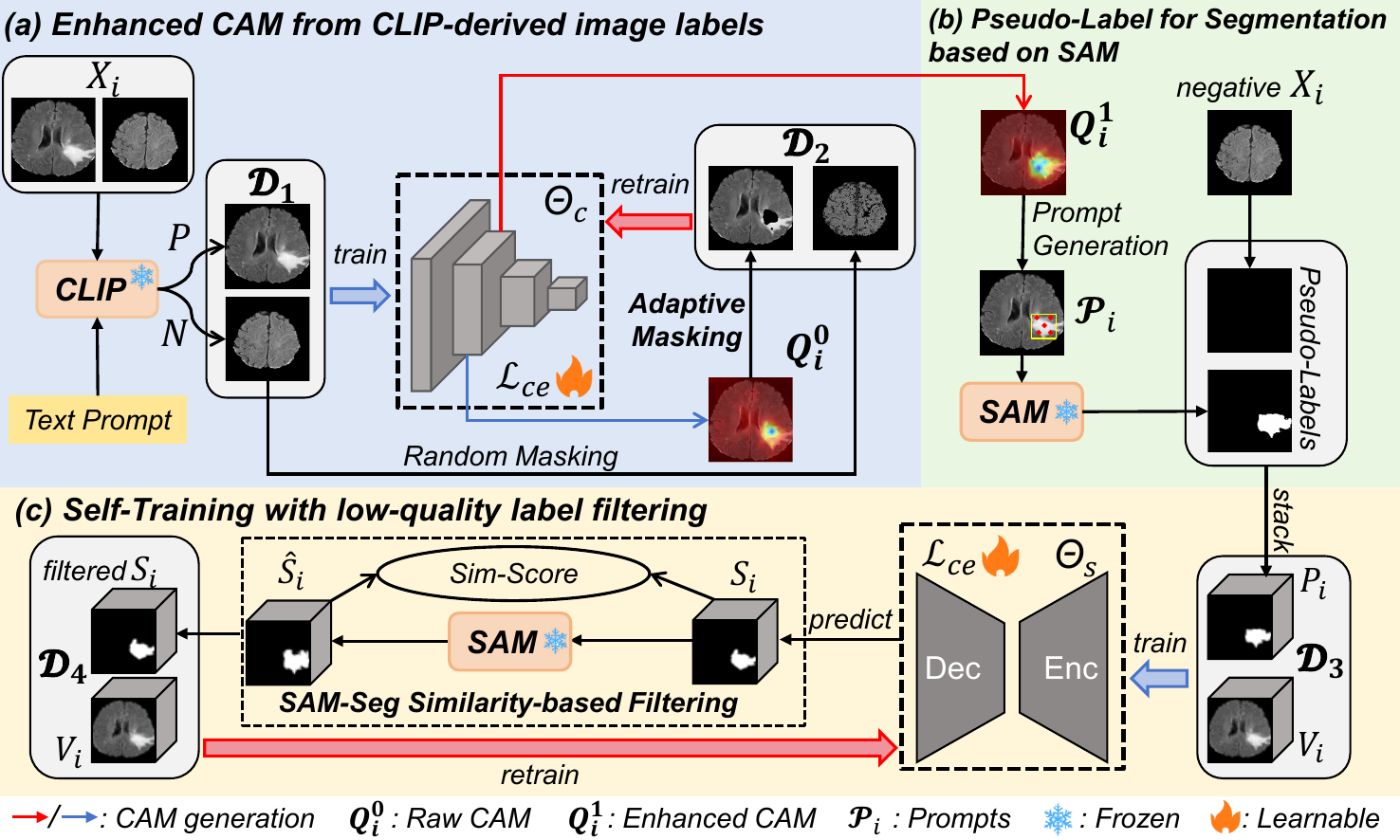}}
\caption{Overview of our proposed CLISC framework for unsupervised brain tumor segmentation.} 
\label{fig1}
\end{figure}

\begin{table*}[t]
    \caption{Quantitative comparison of different unsupervised methods for brain tumor segmentation. The best results are in bold and the second are underlined. $*$ indicates that the p-value $<$ 0.001 in paired t-test compared to the second-best results.}
    \centering
    \resizebox{\textwidth}{!}{
    \begin{tabular}{cccc|ccccc}
\hline
\multirow{2}{*}{Method} & \multirow{2}{*}{Label-free} & \multirow{2}{*}{DSC(\%)} & \multirow{2}{*}{HD$_{95}$(mm)} & \multicolumn{5}{c}{DSC(\%) comparison of different tumor sizes}                             \\ \cline{5-9}
                        &                             &                          &                           & Tiny              & Small           & Medium            & Large          & Huge          \\ \hline
SaLIP~\cite{aleem2024test}                   & \checkmark                           & 56.03±23.09              & 33.17±29.51               & 42.17±28.26         & 49.56±24.29         & 56.64±21.11          & 63.28±18.25         & 67.13±19.62         \\
AMCons~\cite{silva2022constrained}                  & \checkmark                           & 64.37±21.08              & 24.11±17.92               & 51.37±25.93         & 57.22±24.94         & 66.38±22.93          & 68.62±17.34         & 70.1±15.32          \\
GradCAMCons~\cite{silva2022constrained}             & \checkmark                           & 66.01±25.69              & \underline{18.71±16.43}               & 52.23±29.18         & 59.52±27.67         & 64.33±21.76          & 69.14±20.39         & 72.29±13.98         \\
3D IF~\cite{naval2021implicit}                   & \checkmark                           & 67.15±8.89               & 20.64±11.27               & 62.49±10.76         & 65.58±8.86          & 69.32±9.93           & 72.19±8.16          & 73.43±8.27          \\
Sim2Real~\cite{zhang2023self}                & \checkmark                           & \underline{74.86±15.26}              & 24.64±25.95               & \underline{68.51±17.29}         & \underline{72.54±14.97}         & \underline{75.21±12.32 }         & \underline{79.95±10.53}         & \underline{82.76±8.51}          \\
\textbf{CLISC(Ours)}               & \checkmark                           & \textbf{85.60±9.26}$^{*}$      & \textbf{6.72±5.41}$^{*}$        & \textbf{79.67±9.58}$^{*}$ & \textbf{82.49±9.46}$^{*}$ & \textbf{86.10±10.57}$^{*}$ & \textbf{89.39±6.10}$^{*}$ & \textbf{89.93±4.81}$^{*}$ \\ \hline
SAM~\cite{kirillov2023segment}            & \xmark                           & 80.59±9.25               & 9.03±5.12                 & 75.43±10.37         & 78.92±10.19         & 81.36±10.77          & 84.62±8.63          & 86.33±4.58          \\ 
Fullsup  & \xmark                           & 87.97±9.35               & 7.49±13.24                & 77.39±12.71         & 85.54±8.01          & 91.51±5.44           & 91.93±4.14          & 92.78±2.80          \\ \hline
\end{tabular}}
    \label{table1}
\end{table*}

\section{Methodology}
\label{sec:method}
The overall structure of our framework is illustrated in Fig.~\ref{fig1}. First, we use CLIP-derived image-level labels to supervise a classification network and generate enhanced CAM~\cite{jiang2021layercam} with Adaptive Masking-based Data Augmentation (AMDA). Next, segmentation pseudo-labels are obtained by leveraging SAM~\cite{kirillov2023segment} with CAM-derived prompts. Finally, a 3D segmentation network is trained with these pseudo-labels, where low-quality pseudo-labels are filtered out in a self-learning process based on the similarity between the SAM's output and the network's prediction.

\subsection{Enhanced CAM from CLIP-derived image labels}
\label{ssec:2.1}

\subsubsection{CAM from CLIP}
\label{sssec:2.1.1}
CLIP~\cite{radford2021learning} achieves feature alignment by an image encoder $E_{img}$ and a text encoder $E_{text}$. For an input image $X_{i}$, image feature is extracted by $f_{i} = E_{img}(X_{i})$. Text feature is obtained by $g_{c} = E_{text}(\Phi(T_{c}))$, where $\Phi$ indicates the tokenizer and $c$ is class index. The probability of $X_{i}$ belonging to the $c$-th class $p_{i}^{c}$ is obtained by: 
\begin{equation}
    p_{i}^{c} = \frac{e^{Sim({f_{i}}, {g_{c}})}}{ {\textstyle \sum_{j}^{}e^{Sim({f_{i}}, {g_{j}})}} } 
\end{equation}
where $Sim(f_{i}, g_{c})$ is the cosine similarity between $f_{i}$ and $g_{c}$. In this study, text prompts are formulated as $T_{0}$: ``\textit{an image of brain tissue showing typical signal intensity without any regions of abnormal intensity or suspicious mass}" and $T_{1}$: ``\textit{an image of brain tissue showing a tumor with uneven hyperintensity and irregular borders distinct from surroundings}". Let $y_i = argmax(p_i)$ denote the binary pseudo-label for image $X_i$. The entire dataset is denoted as $\mathcal{D}_{1} = \{X_{i}, y_{i}\}_{i = 1}^{N_1}$, where $N_1$ is the number of slices. $\mathcal{D}_1$ is used for training a classification network $\Theta_{c}$ with cross entropy. 

Based on the trained $\Theta_{c}$, we then obtain CAMs that highlight the regions of a training image that contribute most to being classified as having a tumor, and the CAMs can generate prompts for SAM~\cite{kirillov2023segment}. Among various CAM methods~\cite{jiang2021layercam, zhou2016learning, selvaraju2017grad}, we utilized Layer-CAM~\cite{jiang2021layercam}, due to its superiority of providing higher-resolution and more detailed visualizations. Let $Q^0$ denote the raw CAM based on Layer-CAM~\cite{jiang2021layercam}, and it is generated by weighting and combining feature maps from intermediate layers.
\begin{equation}
    Q^{0} = \sum_{k}^{}w_{k}\cdot ReLU(Grad_{k}^{}(c=1) )
    \label{eq1}
\end{equation}
where $w_{k}$ represents the average gradient in the $k$-th channel and $Grad_{k}(c=1)$ denotes the gradient concerning the presence of tumor.
\subsubsection{Adaptive Masking-based Data Augmentation (AMDA)}
\label{sssec:2.1.2}
 As CAM~\cite{jiang2021layercam} usually focuses on the most discriminative part of the image while ignoring other parts of the target, to better mine the entire region of tumors, we propose an AMDA strategy. For each $X_{i}$ with $y_{i}=1$, we identify the most discriminative region $M_{i}$ based on a threshold $T_{\alpha}$ that means the $\alpha$-th percentile in $Q^0$: $M_{i} = Q_{i}^0 > T_{\alpha}$. Moreover, random masks are generated for $X_{i}$ with $y_{i}=0$. Then $X_{i}$  is augmented by masking pixels in $M_i$, and the output is denoted as $\hat{X}_i = X_{i} \cdot (1-M_i)$. The entire dataset is formulated as $\mathcal{D}_{2} = \{\hat{X}_i, y_i\}_{i=1}^{N_1}$, which is used to retrain $\Theta_{c}$. We then apply Layer-CAM again to obtain an updated CAM (denoted as $Q^1$) for each training image with $y_i = 1$.  Note that $Q^1$ mines more margin regions of the tumor, therefore enhancing the quality and accuracy of the CAM-derived ROIs. 


\subsection{Pseudo-Label for Segmentation based on SAM}
\label{ssec:2.2}
SAM~\cite{kirillov2023segment} comprises an image encoder $E_{m}$ and a decoder $D_{m}$ to predict a segmentation mask. Besides, a prompt encoder $E_{prom}$ is designed to align the image and prompt feature. For a given image $X_{i}$, the mask prediction $M'_{i}$ is formulated as:
\begin{equation}
    M'_{i} = D_{m}(E_{m}(X_{i}), E_{prom}(\mathcal{P}_i))
    \label{eq3}
\end{equation}
where $\mathcal{P}_i$ denotes the prompts generated from $Q^1$. We use two types of prompts simultaneously: the bounding box of the ROIs derived from $Q^1$, and the center point (foreground) with four corner points (background) of the bounding box, which effectively clarify tumor regions and noise areas.

For each slice with $y_i =1$, we use Eq~(\ref{eq3}) to obtain the pseudo-segmentation mask, and for $y_i=0$, we set the segmentation mask as empty. These slice-level segmentation masks in a volume are stacked into a 3D pseudo-label $P_{i}$.

\subsection{Self-Training with Low-Quality Label Filtering}
\label{ssec:2.3}
A 3D segmentation network $\Theta_{s}$ is trained based on the above pseudo-labels, and the entire training set is denoted as $\mathcal{D}_3 = \{V_{i}, P_{i}\}_{i=1}^{N_{2}}$, where $N_2$ is the number of volumes. Considering that some pseudo-labels may be noisy due to the accumulated errors in CAM generation and SAM's prediction, we propose a self-training strategy with SAM-Seg Similarity-based Filtering (S3F) of low-quality pseudo-labels.

The self-training involves two rounds of training, and the prediction for $V_i$ from the segmentation model trained in the first round is denoted as $S_i$. Then prompts are generated from $S_{i}$ as the same method as in section 2.2 for SAM~\cite{kirillov2023segment} to obtain its output $\hat{S}_{i}$. Then the training set is updated as $\mathcal{D}_{4} = \{V_i, S_i \medspace |\medspace \mathcal{F}(S_i, \hat{S}_i ) > T_{\beta} \}_{i=1}^{N_2}$, where $\mathcal{F}$ is the similarity score (DSC) with the assumption that a higher consistency between $S_i$ and $\hat{S}_i$ indicates a higher quality of $S_i$, and $T_{\beta}$ is the $\beta$-th percentile of all the $\mathcal{F}$ value in $\mathcal{D}_4$. The $\Theta_{s}$ is retrained with $\mathcal{D}_4$ in the second round, in which low-quality pseudo-labels are filtered out based on SAM-Seg similarity.

\section{Experiments and results}
\label{sec:experiments}

\subsection{Dataset and Implementation}
\label{ssec:3.1}

We evaluated our method on the BraTS2020 that comprises 369 cases each containing 3D volumes in four modalities. In this work, we focus on the segmentation of the whole tumor from FLAIR images and ignore the other modalities. The dataset was split into 7:1:2 for training, validation and testing, respectively. For image preprocessing, we normalized the intensity of the brain area by their mean and std values.

Experiments were conducted in PyTorch on a Linux server with one NVIDIA GeForce GTX 1080Ti GPU. For CLIP~\cite{radford2021learning} and SAM~\cite{kirillov2023segment}, we used the ViT-Base version of the pre-trained model. For the classification network, we utilized the pre-trained ResNet50~\cite{he2016deep} model. The training was conducted using the Adam optimizer, with a learning rate of 0.0001 and a batch size of 8, throughout 200 epochs. For the segmentation network, we utilized a 3D U-Net~\cite{cciccek20163d} with an input size of 128$\times$128$\times$128 and four downsampling and upsampling. The feature channel numbers for the five resolution levels were (16, 32, 64, 128, 256). It was trained using the SGD optimizer, with a learning rate of 0.01 and a batch size of 2, for 400 epochs. For the hyperparameters of AMDA and S3F, we set $\alpha=\beta=20$. As for the evaluation of segmentation performance, volume-level Dice Similarity Score (DSC) and the 95-th percentile of Hausdorff Distance (HD$_{95}$) are utilized.

\subsection{Comparison with Existing Methods}
\label{ssec:3.2}

Our method was compared with five state-of-the-art unsupervised segmentation methods: 
1) SaLIP~\cite{aleem2024test} that utilizes SAM~\cite{kirillov2023segment} to identify ROIs and CLIP~\cite{radford2021learning} to label them, 2) GradCAMCons~\cite{silva2022constrained} that enforces size constraints on Grad-CAM~\cite{selvaraju2017grad} maps, 3) AMCons~\cite{silva2022constrained} that applies entropy regularization on non-weighted activation maps, 4) 3D IF~\cite{naval2021implicit} that uses an auto-decoder feed-forward neural network to learn the distribution of normal tissue and 5) Sim2Real~\cite{zhang2023self} that simulates tumors for training by perturbing 3D polyhedrons. Additionally, we examined two supervised approaches: 1) fully supervised learning (Fullsup) and 2) directly using ground truth to generate bounding box and point prompts for SAM~\cite{kirillov2023segment} for zero-shot inference.

\begin{figure}[t]
  \centering
  \centerline{\includegraphics[width=8.5cm]{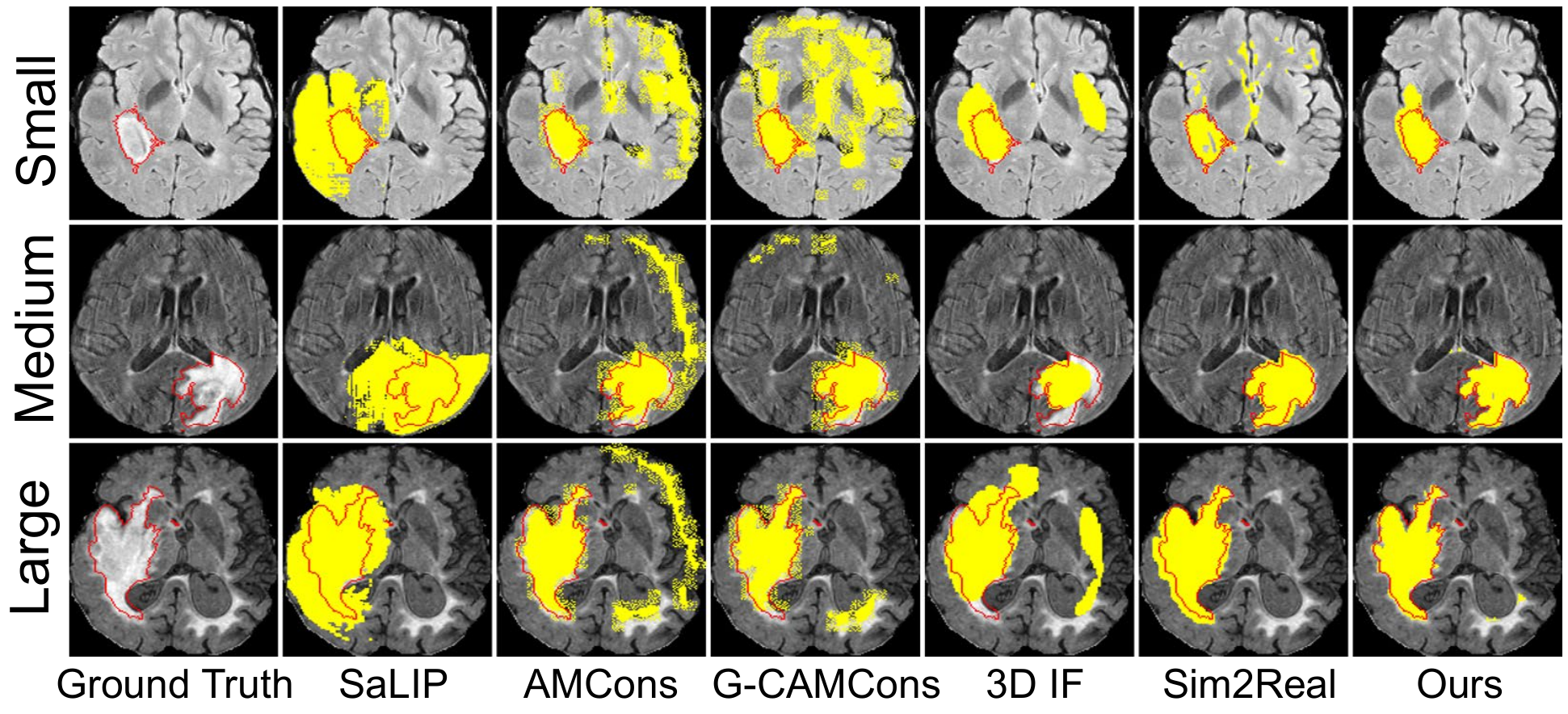}}
\caption{Visual comparison of different unsupervised methods on brain tumor with small, medium, and large sizes. The red curves are contours of the ground truths.} 
\label{fig3}
\end{figure}

The quantitative comparison is presented in Table~\ref{table1}. Our method achieved an average DSC of 85.6\% and HD$_{95}$ of 6.72mm, surpassing the other unsupervised methods by over 10 percentage points on DSC and 10mm on HD$_{95}$. Our method also outperformed SAM~\cite{kirillov2023segment} with prompts generated by ground truth. Though the DSC of our method is slightly lower than that of Fullsup, our method achieved a better HD$_{95}$ value (i.e., 6.72 vs 7.49 mm). To better understand the performance on tumors with different sizes, we split the testing set into five groups with every 20 percentiles of tumor size. Results in Table~\ref{table1} show that our method achieved the best results on all sizes in unsupervised methods. Note that for tiny tumors that are hard to segment, all the other unsupervised methods had poor performance (DSC $<$ 70\%), while our method largely outperformed them, and even obtained better results than Fullsup. Fig.~\ref{fig3} visually compares these methods on tumors of varying sizes, highlighting our method's clear advantages over the other unsupervised approach.





\subsection{Ablation Study}
\label{ssec:3.3}
For ablation study, we first generated CAMs as segmentation using the classification network for testing images, and evaluated the effect of AMDA on the quality of CAMs. Results in Table~\ref{table2} show that AMDA increases the DSC from 51.10\% to 60.22\%. Then we evaluated the performance of SAM on testing images using different prompts generated by our CAM with AMDA. We considered bounding boxes, points, and a combination of both. Results in Table~\ref{table2} demonstrate that a combination of both has obvious advantages, obtaining an average DSC of 74.21\%. Finally, we evaluated the performance of the segmentation model and the impact of S3F. It's shown that training a segmentation model with pseudo-labels by SAM~\cite{kirillov2023segment} increases the DSC to 82.29\% and self-training with S3F further improves it to 85.60\%.

\begin{figure}[t]
    \centering
    \begin{minipage}{0.235\textwidth}
        \centering
        \raisebox{1mm}{\small (a)}  
        \includegraphics[width=\linewidth]{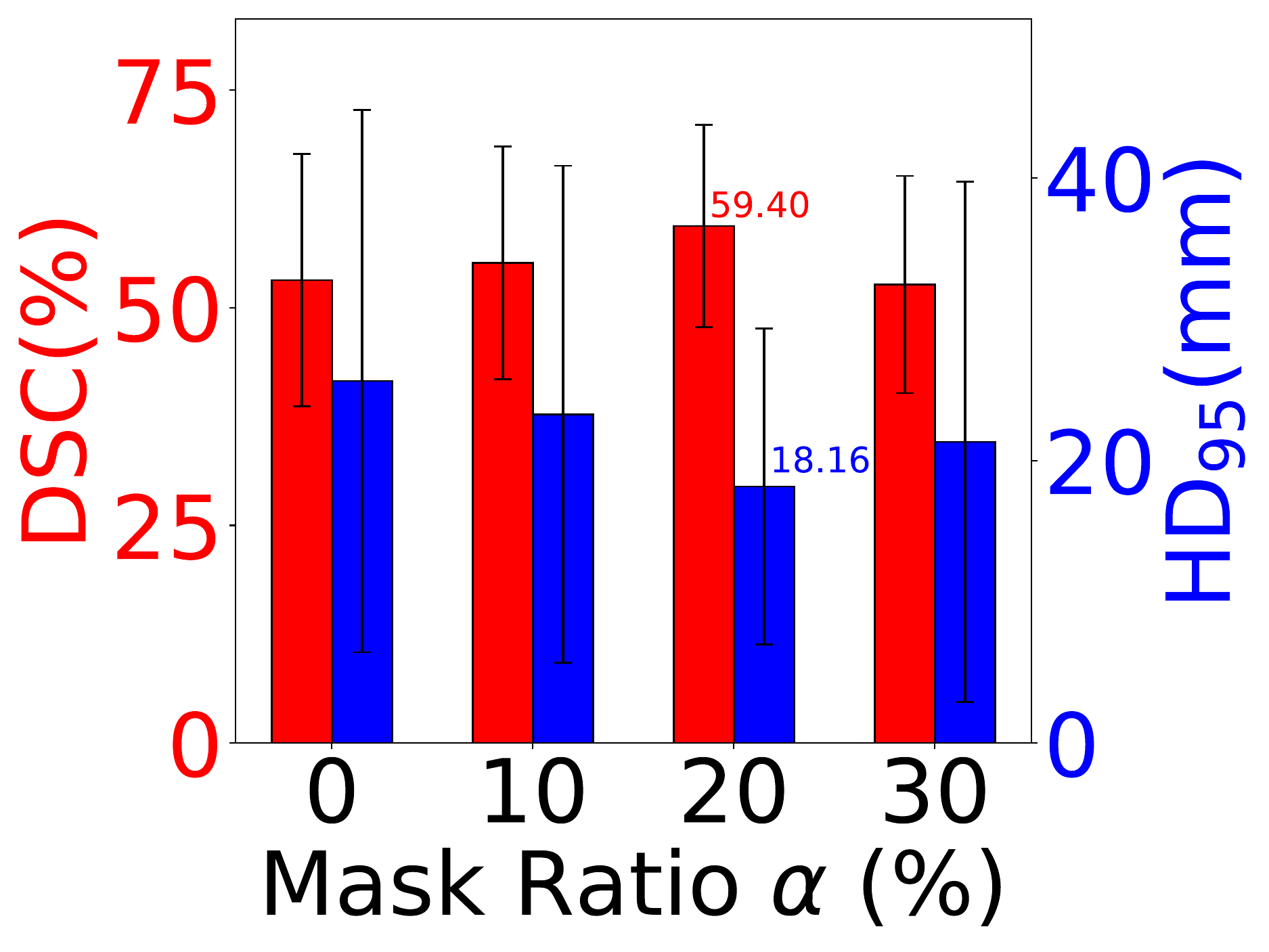}
    \end{minipage}
    \begin{minipage}{0.235\textwidth}
        \centering
        \raisebox{1mm}{\small (b)}  
        \includegraphics[width=\linewidth]{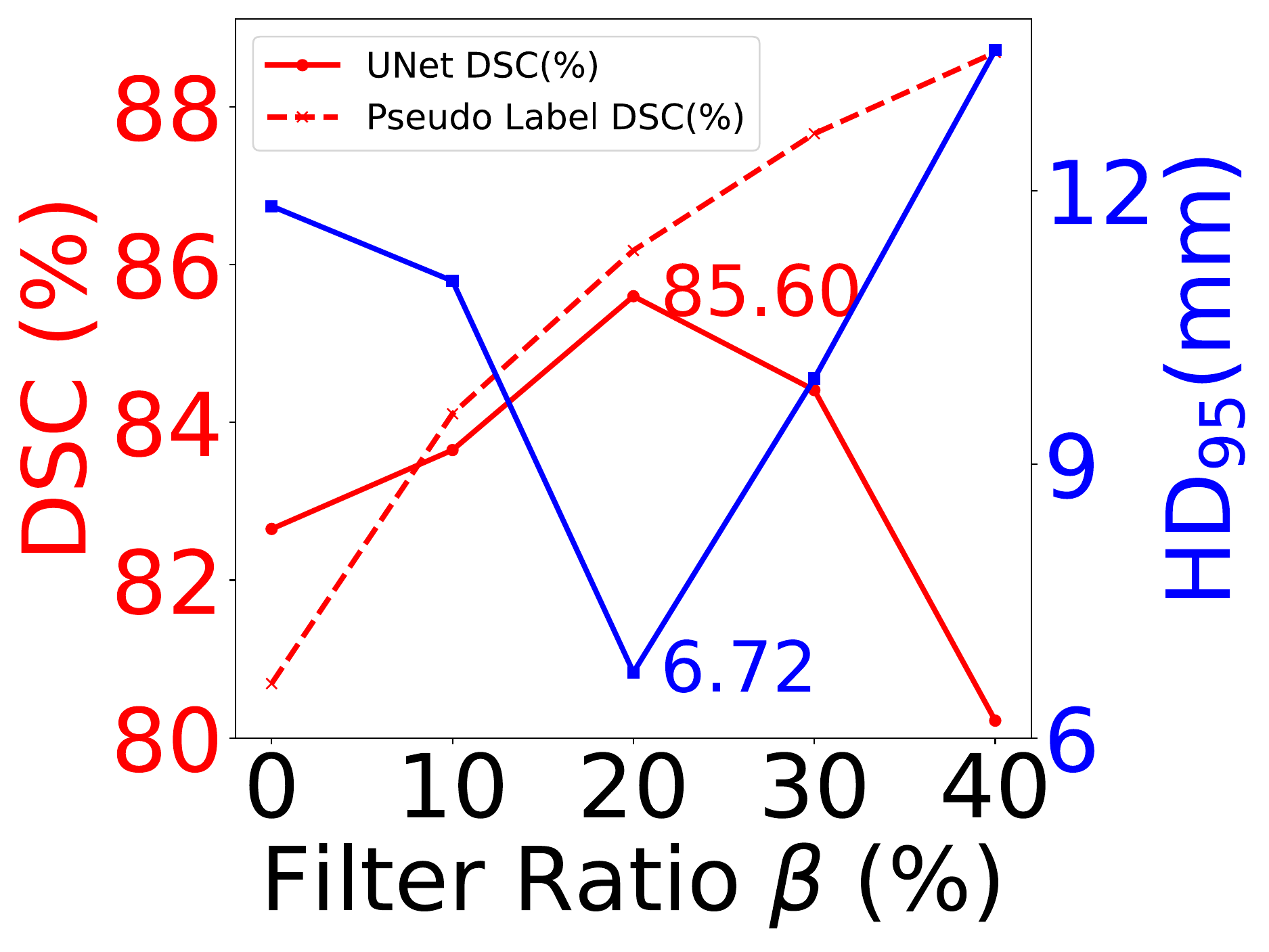}
    \end{minipage}
    \caption{Effect of hyperparameters for AMDA and S3F.}
    \label{fig2}
\end{figure}

We also explored the impact of hyperparameters. For \textbf{AMDA}, we set $\alpha$ to 0, 10, 20 and 30. It's shown in Fig.~\ref{fig2}(a) that $\alpha = 20$ reached the best result. A large $\alpha$ will mask all the tumor areas while the image is still labeled as containing a tumor, which will mislead the model. For \textbf{S3F}, we set $\beta$ to 0, 10, 20, 30 and 40. Fig.~\ref{fig2}(b) shows that as $\beta$ increases, the quality of pseudo-labels gradually improves. However, the number of samples in the training set will be small when $\beta$ is large, leading to weak generalization ability and poor segmentation results. The 3D U-Net~\cite{cciccek20163d} obtains the best performance when $\beta$ is set to 20.

\begin{table}[h]
\caption{Ablation study of our method. Note that all the SAM variants use CAM with AMDA for prompt generation. }
\centering
\renewcommand{\arraystretch}{}
\resizebox{\columnwidth}{!}{
\begin{tabular}{ccc}
\hline
Method                                  & DSC(\%)               & HD$_{95}$(mm)              \\ \hline
raw CAM	                                  &51.10±12.14	       &28.35±21.96 \\
CAM with AMDA  & 60.22±10.78 & 20.10±18.90 \\ \hline
SAM (5 points)                      & 54.30±23.81 & 23.61±19.37 \\ 
SAM (box)                      & 57.76±19.01 & 25.49±23.76 \\
SAM (5 points + box)                      & 74.21±12.95 & 13.46±10.98 \\ \hline
Ours (w/o self-training)                         & 82.29±11.01	& 9.38±9.89 \\
Ours (w/o S3F)                         & 82.01±11.44	& 12.24±16.28 \\
Ours               & \textbf{85.60±9.26}  & \textbf{6.72±5.41}   \\ \hline
\end{tabular}}
\label{table2}
\end{table}


\section{Conclusion}
\label{sec:discussion}
In conclusion, we proposed a novel unsupervised brain tumor segmentation framework CLISC that bridges CLIP and SAM by enhanced CAM. It uses CLIP-derived image labels to supervise a classification network with Adaptive Masking-based Data Augmentation for enhanced CAM. The high-quality CAMs are then used to generate prompts for SAM to obtain segmentation pseudo-labels. Finally, a 3D segmentation network is trained with pseudo-labels in a self-learning process, in which SAM-Seg Similarity-based Filtering is used to select high-quality pseudo-labels. Evaluation on the BraTS2020 dataset shows that our method surpasses five SOTA unsupervised approaches by more than 10 percentage points in terms of DSC. Moreover, it surpasses SAM with prompts from manual labels and matches the performance of fully supervised learning. In the future, we will further investigate this framework's effectiveness in segmenting brain tumor substructures and tumors in other organs.

\section{COMPLIANCE WITH ETHICAL STANDARDS}
\label{sec:5}

This research study was conducted retrospectively using human subject data made available in open access. Ethical approval was not required as confirmed by the license attached
with the open-access data.

\section{ACKNOWLEDGMENT}
\label{sec:6}

This work was supported by the National Natural Science
Foundation of China under Grant 62271115, and by the Fundamental Research Funds for the Central Universities under Grant ZYGX2022YGRH019.

\bibliographystyle{IEEEbib}
\bibliography{refs}

\end{document}